\documentclass{article}

\usepackage{arxiv}

\usepackage[utf8]{inputenc} 
\usepackage[T1]{fontenc}    
\usepackage{hyperref}       
\usepackage{url}            
\usepackage{booktabs}       
\usepackage{amsfonts}       
\usepackage{nicefrac}       
\usepackage{microtype}      
\usepackage{lipsum}		
\usepackage{graphicx}
\usepackage[square,sort,comma,numbers]{natbib}
\usepackage{doi}
\usepackage[utf8]{inputenc}
\usepackage{graphicx}
\usepackage{float}
\usepackage{amsmath}
\usepackage{array}
\usepackage{titlesec}

\title{Harfang3D Dog-Fight Sandbox: A Reinforcement Learning Research Platform for the Customized Control Tasks of Fighter Aircrafts}

\author{ {\hspace{1mm}Muhammed Murat Özbek}\\
	Istanbul Technical University\\
    ozbekm17@itu.edu.tr\\
	\And
	{\hspace{1mm}Süleyman Yıldırım} \\
	Koç University\\
	suleymanyildirim22@ku.edu.tr
	\And
	{\hspace{1mm}Muhammet Aksoy} \\
	Istanbul Technical University\\
	aksoym15@itu.edu.tr\\
	\And
	{\hspace{1mm}Eric Kernin} \\
	HARFANG 3D NWNC\\
    eric.kernin@harfang3d.com\\
	\And
	{\hspace{1mm}Emre Koyuncu} \\
	Istanbul Technical University\\
    emre.koyuncu@itu.edu.tr
}

\date{}

\renewcommand{\headeright}{}
\renewcommand{\undertitle}{}
\renewcommand{\shorttitle}{}

\hypersetup{
pdftitle={Harfang3D Dog-Fight Sandbox: A Reinforcement Learning Research Platform for the Customized Control Tasks of Fighter Aircrafts}
}

\begin{document}
\maketitle

\begin{abstract}
{\it Abstract - The advent of deep learning (DL) gave rise to significant breakthroughs in Reinforcement Learning (RL) research. Deep Reinforcement Learning (DRL) algorithms have reached super-human level skills when applied to vision-based control problems as such in Atari 2600 games where environment states were extracted from pixel information. Unfortunately, these environments are far from being applicable to highly dynamic and complex real-world tasks as in autonomous control of a fighter aircraft since these environments only involve 2D representation of a visual world. Here, we present a semi-realistic flight simulation environment Harfang3D Dog-Fight Sandbox for fighter aircrafts. It is aimed to be a flexible toolbox for the investigation of main challenges in aviation studies using Reinforcement Learning. The program provides easy access to flight dynamics model, environment states, and aerodynamics of the plane enabling user to customize any specific task in order to build intelligent decision making (control) systems via RL. The software also allows deployment of bot aircrafts and development of multi-agent tasks. This way, multiple groups of aircrafts can be configured to be competitive or cooperative agents to perform complicated tasks including “Dog Fight”. During the experiments, we carried out training for two different scenarios: navigating to a designated location and within visual range (WVR) combat, shortly “Dog Fight”. Using Deep Reinforcement Learning techniques for both scenarios, we were able to train competent agents that exhibit human-like behaviours. Based on this results, it is confirmed that Harfang3D Dog-Fight Sandbox can be utilized as a 3D realistic RL research platform.
}
\end{abstract}

\section{Introduction}
The training of machine learning models to make a sequence of decisions is known as Reinforcement  Learning. In an uncertain and potentially complex environment, the RL agent meets a game-like simulation and learns to achieve a specified goal. In order to accomplish a given task, the agent uses trial and error method. Consequently, it is given either rewards or penalties for the actions that it takes. As a result, starting with an agent that behaves completely randomly, the algorithm is capable of developing sophisticated tactics without any prior knowledge about the environment or the game. The ability to adapt stochastic, complex, and dynamics environments to perform any task with superhuman skills makes Reinforcement Learning a powerful machine learning tool in the area of robotics and control. Therefore, researchers have been using RL in various control tasks \cite{birinci}, \cite{iki}, \cite{uc}, \cite{dort}: manipulation \cite{bes}, \cite{alti}, \cite{yedi}, \cite{sekiz}, locomotion \cite{dokuz}, \cite{on}, navigation \cite{onbir}, \cite{oniki},\cite{onuc}, \cite{ondort}, flight \cite{onbes}, \cite{onalti}, interaction \cite{onyedi}, \cite{onsekiz}, motion planning \cite{ondokuz}, \cite{yirmi} and more. 

However, training of RL models differs from traditional machine learning algorithms due to its game-like nature. Unlike supervised and unsupervised learning, there is no labeled data nor a static dataset in Reinforcement Learning. This means that dataset changes dynamically as the agent interacts with its environment because the training samples comes from agent’s previous experiences. This poses a number of problems when it comes to training an RL agent, which are high sample complexity, training time and realistic environment. Training agents in actual robotic systems is not a feasible option because of replicability, financial, time, and safety concerns. Hence, simulation environments customized for specific tasks are traditionally used for training and experimentation in Reinforcement Learning research \cite{bes}, \cite{dokuz}, \cite{yirmibir}. On top of that, distributional training methods have been utilized to increase the sampling diversity and decrease the training time \cite{yirmiiki},\cite{yirmiuc}, \cite{yirmidort},\cite{yirmibes}. A downside of carrying out experiments in a simulated environment is transferring the model into the real world. Researchers proposed several methods to overcome this domain problem, which is brought about by the lack of fidelity and stochasticity in the simulated environments, including calibration \cite{alti}, \cite{yirmialti}, utilizing real-world data samples \cite{onuc}, \cite{otuz}, \cite{otuzbir}, and domain randomization \cite{onalti}, \cite{yirmiyedi}, \cite{yirmisekiz}, \cite{yirmidokuz}. But still, how realistic a simulation environment is an important factor to be able to solve this problem.

Grounded on the facts explained above, it can be concluded that having access to a realistic, customizable and flexible simulation software that is specialized in the desired field is crucial to experiment with RL. To address the need of such platform in the area of autonomous control of fighter aircrafts, we propose the RL experimentation software Harfang3D Dog-Fight Sandbox. Harfang3D Dog-Fight Sandbox simulates an air-to-air confrontation scene with continuous sensory measurements received from the environment (e.g., Euler angles, coordinates, velocity vectors, and acceleration). Through the proposed platform, one can experiment with different simulation parameters, customize various tasks for fighter aircrafts, build multi agent systems, and run multiple parallel sessions during training by making use of the distributed structure. Moreover, the software can also be interfaced with the OpenAI Gym library \cite{yirmibir}. For experimentation purposes, we trained our model using the Twin Delayed Deep Deterministic Policy Gradient (TD3) algorithm \cite{otuziki} without resorting to any real-world data, or expert knowledge. 

\section{Harfang3D Dog-Fight Sandbox: Reinforcement Learning Research Platform}

Harfang3D Dog-Fight Sandbox is an easy-to-adapt, cross-platform, multi-language, powerful and optimized solution to integrate with embedded systems, into existing environments and combining features to meet both the general public and military challenges of real-time 3D imaging: providing strategic autonomy, reactivity, flexibility, interoperability and durability. It is entirely written in C++, with high level language API in Python, Lua and Golang. 

\begin{figure}[ht] 
    \centering
    \includegraphics{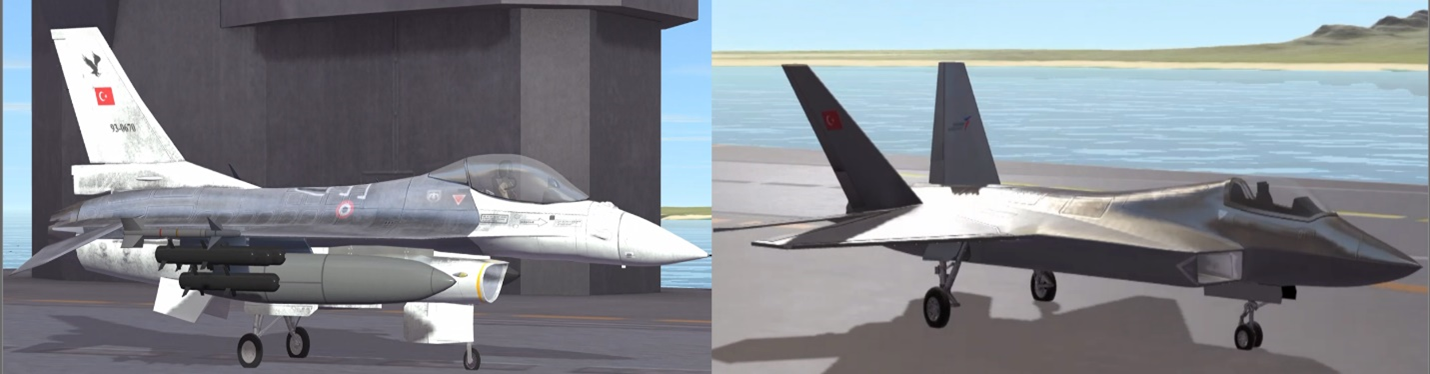}
    \caption{Some examples of the realistic renders of the fighter jets. F-16 jet is on the left side, and TF-X is on the right side.}
\end{figure}

The platform provides features that facilitate adaptable, flexible, and inclusive research environment. The main features of Harfang3D Dog-Fight Sandbox include different control modes, custom scenarios,  custom physics models, distributed run, and renderless mode eliminating the need of using a graphical interface.

\subsection{Runtime}

\subsubsection{Synchronization Modes}
Harfang3D Dog-Fight Sandbox is capable of running above 1000 FPS on a modern PC if needed. But, this number may vary depending on the simulation run mode. We can investigate the simulation run modes in two parts as synchronous and asynchronous modes. In synchronous mode, simulation and DL model have to run at the same frequency because simulation is not updated unless the model gives a command as they run consecutively. Therefore, synchronous mode limits the frame rate of the simulation to approximately 50 Hz. However, simulation can run much faster in asynchronous mode because in this mode, both processes run independently at their own pace without having to wait for each other's response. For instance, if the simulation runs at 1000 Hz and the DL model runs at 50 Hz, the model is going to make 1 inference at every 20 simulation steps since they can run simultaneously. In most cases, real-life scenarios can be modeled more accurately in asynchronous mode. Still, to avoid temporal problems, synchronous mode is preferred for experimentation (OpenAI games and applications run in synchronous mode) due to possible discrepancies between simulation frame rate and algorithm speed. Since simulation and algorithm run consecutively at each step, any temporal constraint is automatically eliminated.

\subsubsection{Distributed Run}
One of the best properties of Harfang3D Dog-Fight Sandbox is that agents can independently run in different environments in parallel at the same time. Additionally, multiple agents can also run in one environment as cooperative or competitive groups. This will give us the opportunity to decrease training time and to boost exploration as well as to create highly complex scenarios and complicated tasks.

For example, this property may be exploited to speed up the training process by running one agent in multiple sandboxes. This way, one agent can take advantage of numerous training samples coming from variety of sources. As this may multiply the computational cost, it will also immensely accelerate the training while enforcing exploration.

Another example is to train multiple agents in multiple environments in parallel. With this setup, different hyperparameter configurations can be experimented at the same time in parallel when training your agent. After that, the best agent can be trained optimally in multiple environments as discussed in the first example.

\begin{figure}[ht] 
    \centering
    \includegraphics{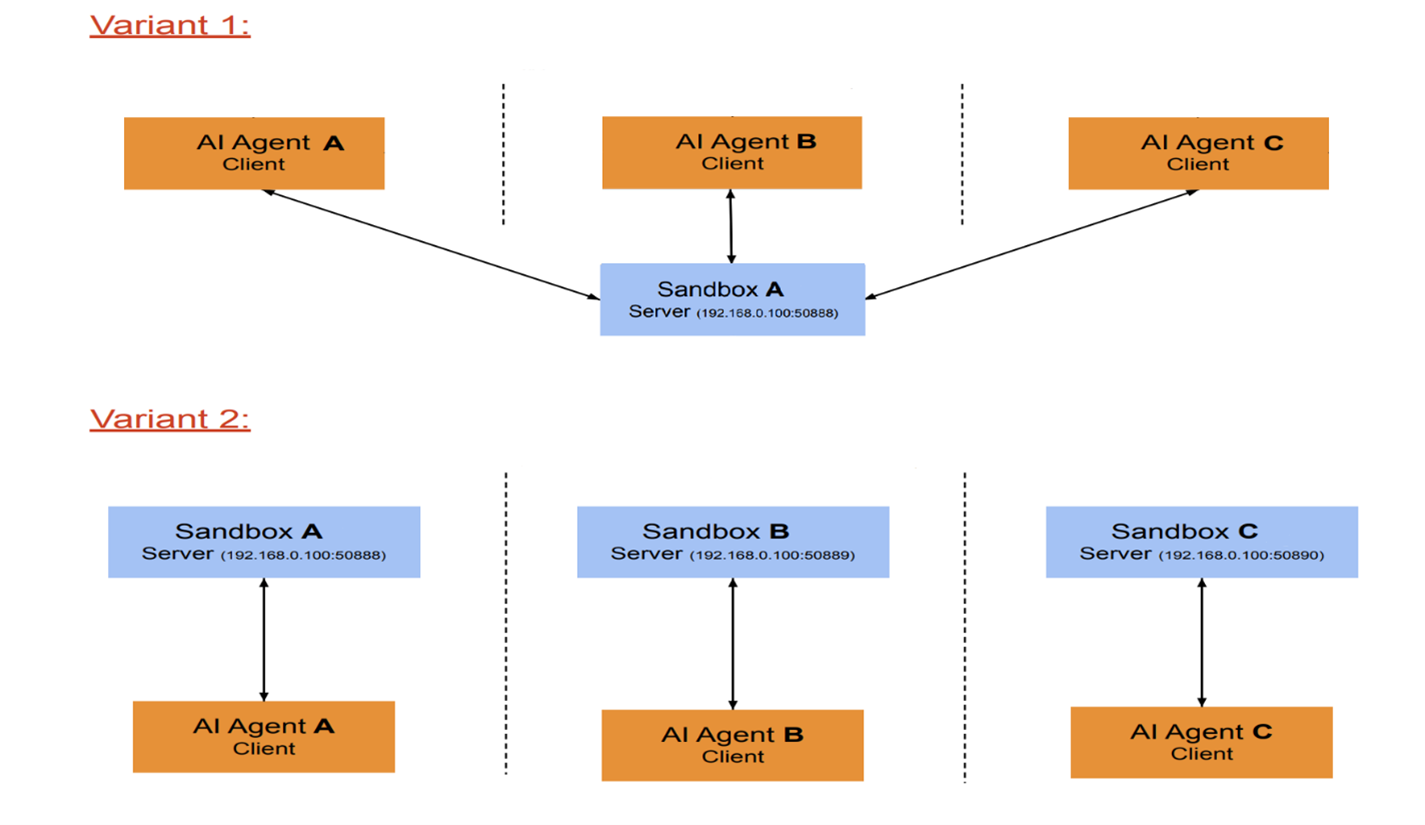}
    \caption{Distributional architecture of Harfang3D Dog-Fight Sandbox and possible parallel running scenarios.}
\end{figure}

\subsection{Simulation Components}
Harfang has mainly two different trainable simulation components. These are aircrafts and missiles. There are five models of fighter jets including world-renowned F-16 and recently being developed TFX. Missiles, on the other hand, consist of two categories (SAM and AAM) and six models in total. Each of these units have unique specialties due to distinct characteristics.

\begin{figure}[ht] 
    \centering
    \includegraphics[scale=0.35]{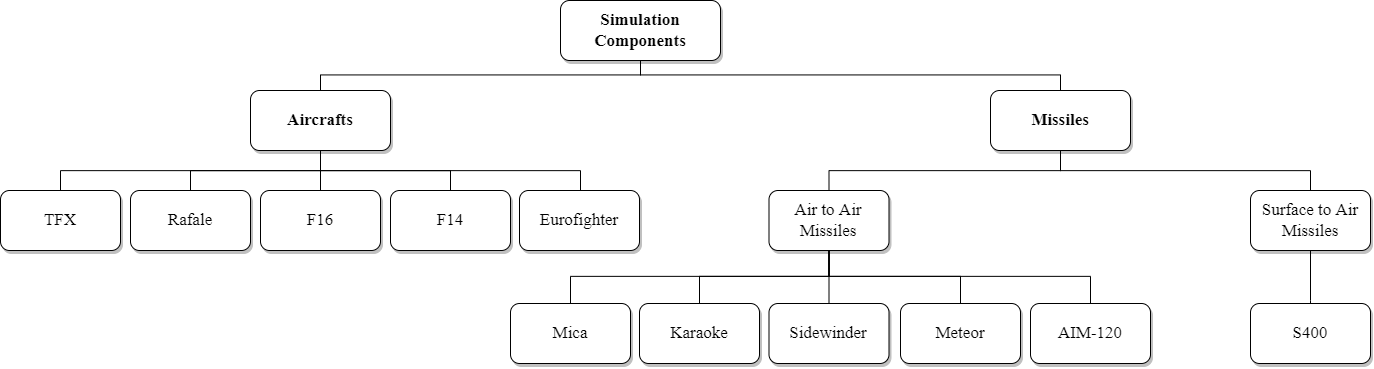}
    \caption{A general diagram for the simulation components and their sub categories.}
    \label{fig:diag1}
\end{figure}

\subsubsection{Aircrafts}
There currently are five different fighter aircrafts in the simulation such as Turkish Fighter X (TF-X) by Turkish Aerospace Industries (TAI), two variants of F-14 by Northrop Grumman (USA), F-16 by General Dynamics (USA), and Rafale by Dassault (France). Model parameters of the aircrafts are listed in Table \ref{tablo1} and they can be altered as desired. Due to the fact that each aircraft has distinct flight dynamics model, each of them has advantageous and disadvantageous aspects accordingly.
Therefore, during training, AI agents experience different aircraft with varying dynamics and ammunition configurations. Consequently, agents are simulated against various opponents and get to learn complex tactics. There are two types of weapons: missiles and machine guns. Different missile configurations of each aircraft can also be found in Table \ref{tablo1}. Prior to the launching of a missile, the target must be locked. The ammunition of the machine gun is not limited.

\begin{table}[ht]
\centering
\renewcommand\arraystretch{1.2}
\begin{tabular}[t]{lccc>{\raggedright}p{0.1\linewidth}>{\raggedright\arraybackslash}p{0.1\linewidth}}
\toprule
& TFX & Rafale & F16 & F14 & Eurofighter \\
\midrule
Thrust Force &20&15&15&10&13\\
Post Combustion Force  &20&7.5&15&5&9\\
Angular Frictions  &0.000175&0.000165&0.000175&0.000175&0.00019\\
Speed Ceiling Force  &2500&2200&1750&1750&2500\\
Max Safe Altitude &25000&15240&15700&15700&16800\\
Max Altitude &30000&25240&25700&25700&26800\\
Missile Number  &4&6&12&4&6\\
Missile Config &4xAIM-120&2xMica 4xMeteor&8xKaraoke  2xAIM-120 2xCFT & 4xSidewinder&2x Meteor     4xMica\\
\bottomrule
\end{tabular}
\caption{This table demonstrates the different parameters of five different fighter aircrafts and their configurations.}
\label{tablo1}
\end{table}%

\subsubsection{Missiles}
In this simulation, missiles can be gathered under two categories. These are air-to-air missiles (AAMs) and surface-to-air missiles (SAMs). Mica, Karaoke, Sidewinder, Meteor, and AIM-120 are currently available air-to-air missiles in Harfang. For the surface-to-air missile category, S-400 is a popular option that is provided in this simulation. Properties of each missile differ from each other to enrich customizability.

\begin{table}[ht]
\centering
\renewcommand\arraystretch{1.2}
\begin{tabular}[t]{lcccc>{\raggedright}p{0.1\linewidth}>{\raggedright\arraybackslash}p{0.1\linewidth}}
\toprule
&Mica & Karaoke & Sidewinder & Meteor& AIM-120 & S-400\\
\midrule
Thrust Force &150&70&100&80&120&200\\
Endurance  &15&35&20&40&20&210\\
Damage  &20-30&50-70&30-40&40-60&25-35&100\\
Angular Frictions  &0.00014&0.00005&0.00008&0.00005&0.00008&0.000025\\
\bottomrule
\end{tabular}
\caption{This table demonstrates the general properties of each missile. Damage values are given as an interval rather than a strict value because the magnitude of the damage is chosen randomly from the given value range. S-400 missile is considered as the most powerful missile in this simulation; thus, its damage is directly set to 100 and it can destroy an airplane with a single hit.}
\label{tablo2}
\end{table}%

\paragraph{Air to Air Missiles}
AAMs can be differentiated according to four main properties: thrust, endurance, damage, angular friction. Mica has the most thrust power, which makes it the fastest missile among four others listed in Table \ref{tablo2}, but it also has the least endurance. Karaoke gives the largest amount of damage even-though it is the slowest missile, while Meteor has the longest endurance. Table \ref{tablo2} demonstrates the properties of five different missile variants. Please note that these values can be easily varied.

\paragraph{Surface to Air Missiles}
One of the most popular surface-to-air missile system is S-400. This system has the longest endurance and the most thrust power when compared to its counterparts. Therefore, S-400 is used in the Harfang 3D Dog Fight sandbox. Table \ref{tablo2} shows physical model parameters of S-400.

\subsection{Scenario Options}
Environment customizability is a key property in RL research. For this purpose, Harfang3D Dog-Fight Sandbox possesses an highly adaptable architecture which enables the creation of diverse tasks and scenarios from many perspectives. This includes being able to deploy different number of players, appoint other bot, AI, or human players as enemy or ally, program custom physical models, modify physical aspects (e.g., wing area) or other features of an aircraft. When combined with its diverse, large, and continuous observation space, it makes Harfang3D Dog-Fight Sandbox a powerful tool capable of creating numerous distinct scenarios. The concept is straightforward: In a three-dimensional space, aircrafts will fight each other. The starting point is always an aircraft carrier. Different protagonists can be piloted by a human, by an autopilot or by an AI that has been developed outside the Sandbox.

\begin{figure}[ht] 
    \centering
    \includegraphics[scale=0.35]{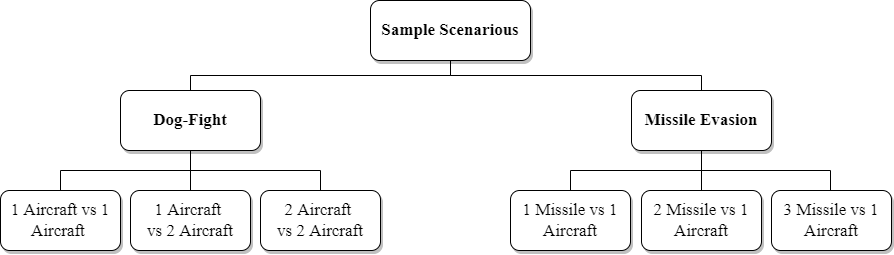}
    \caption{This diagram illustrates sample customized scenarios that can be implemented in Harfang.}
\end{figure}

We have established two main use cases for Harfang: missile evasion and dog fight. Dog fight can be performed in 1 versus 1, 1 versus 2, and 2 versus 2 format. Multiple agents from the same team can cooperate while communicating with each other to defeat the opponent team. On the other hand, missile evasion can be performed in 1 versus 1, 1 versus 2, and 1 versus 3 format. For now, simulation only supports a maximum of three missiles at once for an aircraft.

\subsubsection{Autopilot Mode}
Prior to handing over the controls to an AI, it is important that the aircraft can be steered by an autopilot. The latter simply operates the heading and altitude of the plane, as well as the thrust. Once the autopilot is capable of steering the aircraft, the interfacing with an external AI is no longer an issue. In the abscence of an external AI control, the algorithm provided with the Sandbox to simulate combat scenarios is fairly elementary:
\begin{enumerate}
    \item Head to the target.
    \item Determine the altitude of the target.
    \item If the altitude is below a certain limit, climb to a certain altitude.
    \item If the target is in range and aligned, fire the machine gun.
    \item If the target is locked, fire a missile.
    \item Wait between two missile shots (to avoid unloading everything at once).
\end{enumerate}

\subsection{Physics Model}
In an air combat scenario, the physics of the aircraft is critical. The model used in the Sandbox is a compromise between simplicity and efficiency.

\subsubsection{Air Density}
Air density is an important parameter of the physical model because thrust power that is generated by the turbines is strongly dependant on air density.Air density is represented as $\rho$. For calculations, ideal gas constant R, molar mass of dry air M, and gravitational acceleration G are accepted as 8.3144621 J/($mol.k$), 0.0289652 $kg/mol$, 9.80665 $m/s^2$, respectively. When the necessary calculations are done, it can be seen that the air density decreases as the altitude increases.Temperature is accepted absolute temperature.Note that temperature is kept constant throughout the simulation but it can also be varied.

\begin{equation} 
\rho = \frac{p}{RT}
\end{equation}
 
\subsubsection{Velocity and Dynamic Pressure}
The aircraft responses depend essentially on its velocity, i.e. on the pressure applied by the atmosphere on the wing (and other control surfaces). In the aerodynamic model of the Sandbox, the dynamic pressure is evaluated first in order to obtain a coefficient which will be used in several other parameters calculations.

\begin{equation} 
q = \frac{\rho v^2}{2}
\end{equation}

Where q is dynamic pressure.$\rho$ is density of fluid and v is velocity.For the dynamic pressure in 3 axes, x, y and z components are calculated separately.

\subsubsection{Forces}
Several forces come into play in the flight simulation: gravity, lift, drag, turbine thrust. Lift and Drag will vary with dynamic pressure. The wind force is not taken into account.

\begin{figure}[ht] 
    \centering
    \includegraphics[scale=0.4]{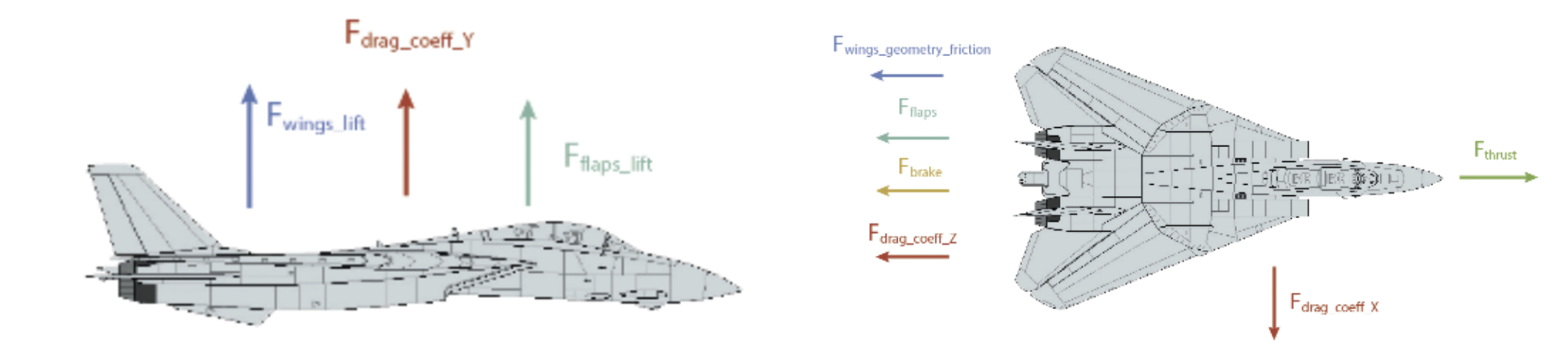}
    \caption{Aerodynamic forces acting on the aircraft model.}
\end{figure}

\begin{equation} 
\vec{F}_{lift} = q_{z}.(\vec{F}_{wings{\_}lift} + \vec{F}_{flaps{\_}lift}) 
\end{equation}

\begin{equation} 
\vec{F}_{drag} = \frac{\vec{V}}{||\vec{V}||}.\vec{q}_{z}.(\vec{F}_{drag{\_}coeff} + \vec{F}_{flaps}+ \vec{F}_{wings{\_}geometry{\_}friction}) 
\end{equation}

\begin{equation} 
\vec{F}_{move} = \vec{F}_{thrust} + \vec{F}_{lift} - \vec{F}_{drag} + \vec{F}_{gravity}
\end{equation}

\subsubsection{Moments}
The pilot, by operating the controls, acts on the moments (angular forces) of the aircraft: pitch, yaw and roll. The rotational movement of the aircraft is the sum of these three moments.
\begin{figure}[ht] 
    \centering
    \includegraphics[scale=0.5]{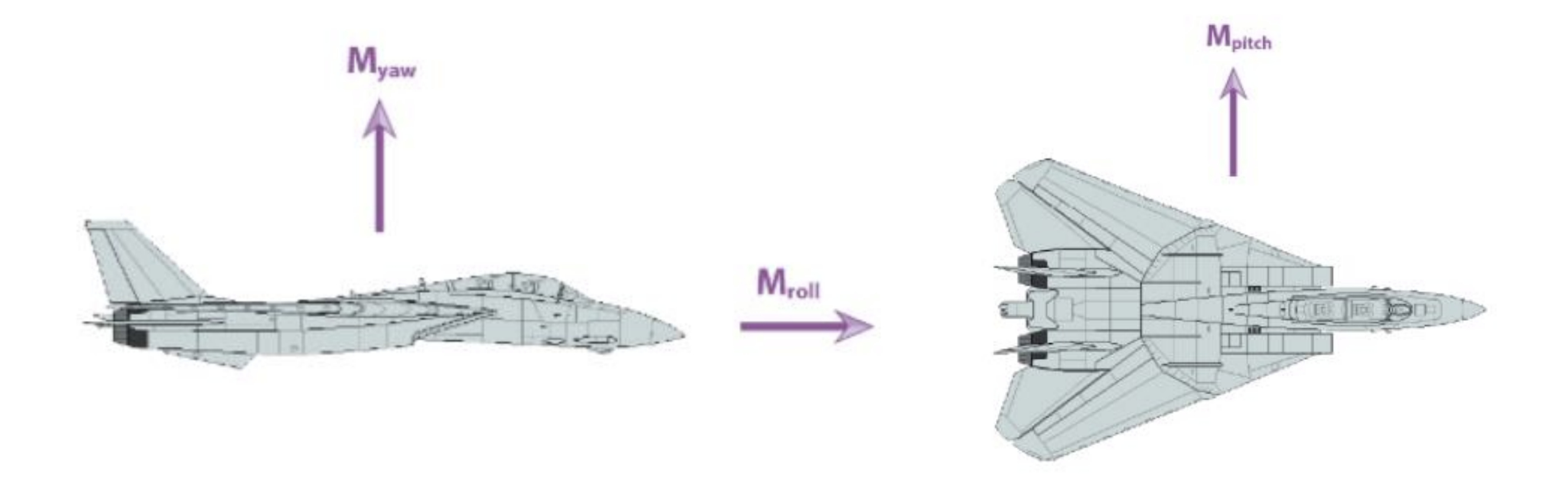}
    \caption{Aerodynamic moments acting on the aircraft model.}
    \label{fig:result}
\end{figure}

\begin{equation} 
\vec{M}_{pitch} = \vec{X}_{axis}.q_{z}.(pitch{\_}Level).( pitch{\_}friction) 
\end{equation}

\begin{equation} 
\vec{M}_{yaw} = \vec{Y}_{axis}.q_{z}.(yaw{\_}Level).( y aw{\_}friction) 
\end{equation}

\begin{equation} 
\vec{M}_{roll} = \vec{Z}_{axis}.q_{z}.(roll{\_}Level).( roll{\_}friction) 
\end{equation}

Total moment can be found from;

\begin{equation} 
\vec{M} = \vec{M}_{pitch} + \vec{M}_{roll} + \vec{M}_{yaw}
\end{equation}

Where $q_z$ is the local Z component of dynamic pressure vector. The Z axis is the front component of plane displacement. If $q_z = 0$, that means the plane only moves along its vertical (Y) and/or lateral (X) axis, the aircraft is in a stall and can no longer be controlled. If $q_z = 1$, that means the plane is moving along its frontal (Z) axis, the aircraft is in stable flight.

\subsubsection{Customized Physics}
\renewcommand{\arraystretch}{2}
\setlength{\arraycolsep}{12pt}
\[
	\left[
     	\begin{array}{ccc}
     	1 & tan(\theta)sin(\phi) & cos(\phi)tan(\theta) \\
	    0 & cos(\phi) & -sin(\phi) \\
	    0 & sec(\theta)sin(\phi) & sec(\theta)cos(\phi) \\
	    x & y & z \\
    	\end{array}
	\right]
\]

The sandbox currently runs on basic physics dynamics; nevertheless, users can easily configure customized dynamics for either aircrafts or missiles. Using the function $set\_machine\_custom\_physics\_mode$, user can choose which aircraft or missile is to be modified. Then, the aerodynamics matrix demonstrated above can be utilized to transfer custom dynamics model into the simulation. Here, $x$, $y$ and $z$ represent the Cartesian coordinates of the plane while $\theta$ and $\phi$ angles represent the data which come from the dynamics. After that, with the help of $update\_machine\_kinetics$ function, position data from the dynamics function can be sent to sandbox for update at each time frame.

\subsection{Graphics and Environment}
The visual realism of the simulation is a key factor in creating an immersive experience for the Sandbox user, especially when using a virtual reality headset. The purpose of the Sandbox is multifaceted, but since one of the objectives is to confront human pilots with algorithm-driven aircraft, the quality of the immersion is central in the acceptance of the simulation.

\subsubsection{Rendering Pipeline}
The Sandbox uses a rendering pipeline that aims to provide a rich and complex visualisation without compromising the performance of an application that is implemented in the Python language.

The rendering is split into several passes that are calculated sequentially as follows, for each frame:
\begin{enumerate}
    \item Rendering of water reflections
    \item Rendering of the ocean and sky
    \item Rendering of the 3D scene's main elements
\end{enumerate}

On top of these 3 passes, two layers overlay 2D information about the status of the simulation as well as 3D information regarding the debugging process.

\begin{figure}[ht] 
    \centering
    \includegraphics[scale=0.2]{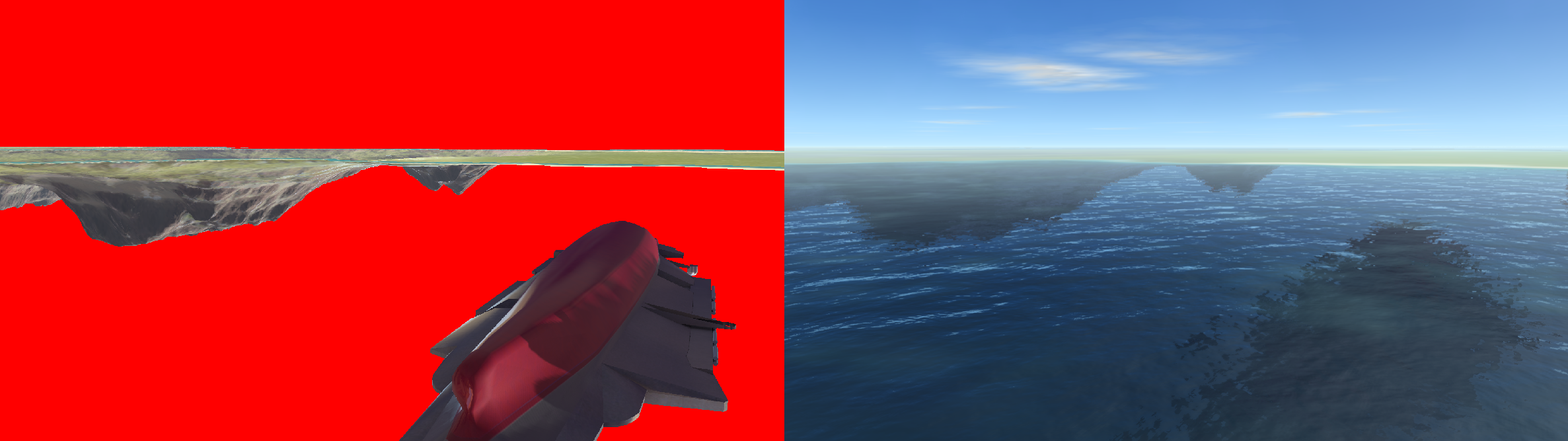}
    \caption{These images depict how the reflections on the ocean is rendered. Left image shows the scene without the ocean and right image shows the scene after the ocean is added.}
    \label{fig:ray1}
\end{figure}

\paragraph{Reflection Rendering Pass}
The scene is rendered from a camera that is mirrored to the main view angle, relative to the plane of the ocean. All solid 3D objects are included in this pass, be they static or dynamic (terrain, ships, aircraft), okyanus uzerindeki yansimalar \ref{fig:ray1} gorselinde detayli gorunebilir.

\paragraph{Ocean and Sky Rendering Pass}
The ocean and the sky are rendered by a fairly simple raytracing routine included in a GPU shader. The framebuffer of the reflection in the water, obtained from the previous pass, is used for that purpose.
The surface of the ocean is a 3D plane, which can intersect with 3D models of the scene (ship hulls, aircraft grounded in the water, emerged terrain). This is why the DepthBuffer is also generated in this raytracing calculation.
\\
\\
Our commitment to provide a solution that can be accessed on standard hardware specifications led us to avoid using RTX extensions.

\paragraph{3D Scene Rendering Pass}
The 3D scene that is the main part of the simulation (aircraft, vehicles, weapons, terrain and particles) is rendered through the primary camera ( subjective view camera, or any other external view of the aircraft).
This scene is rendered over the ocean and the sky. As the ocean DepthBuffer has been rendered, the parts of the 3D models below the water surface are not drawn. Passes 2 and 3 thereby combine seamlessly to create a coherent image.

\begin{figure}[ht] 
    \centering
    \includegraphics[scale=0.2]{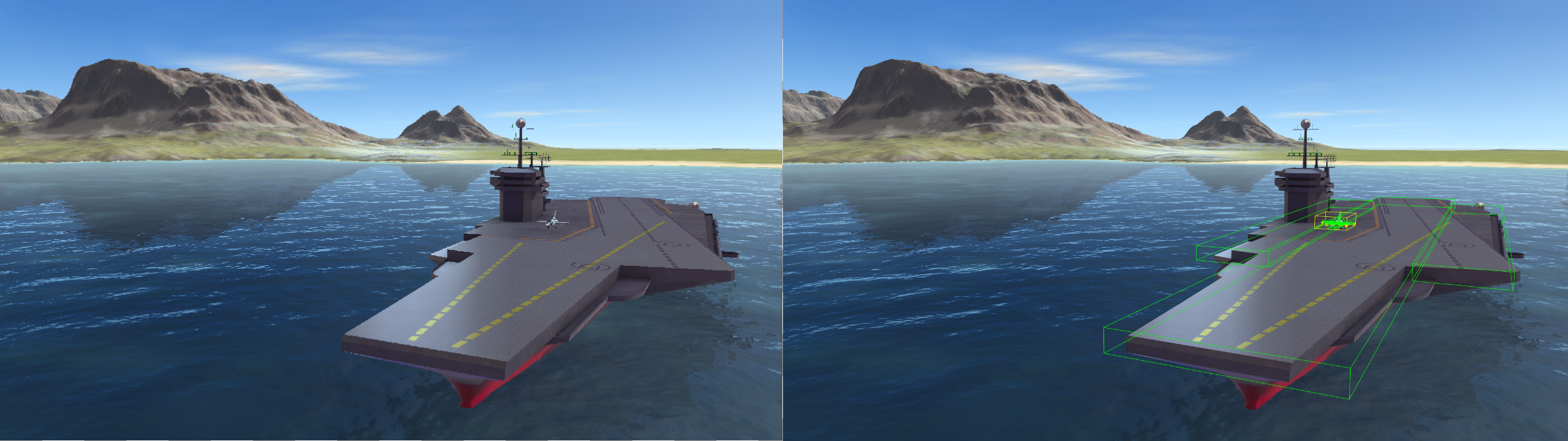}
    \caption{Three dimensional objects are created independently before being integrated. You can see the example of this aircraft carrier.}
    \label{fig:ray2}
\end{figure}

\paragraph{3D Overlays Pass}
The 3D texts as well as the debug displays (3D vectors, flight paths, collision boxes) are rendered in this pass. You can see the generation of the collision boxes as an example in Figure \ref{fig:ray2}.

\begin{figure}[ht] 
    \centering
    \includegraphics[scale=0.4]{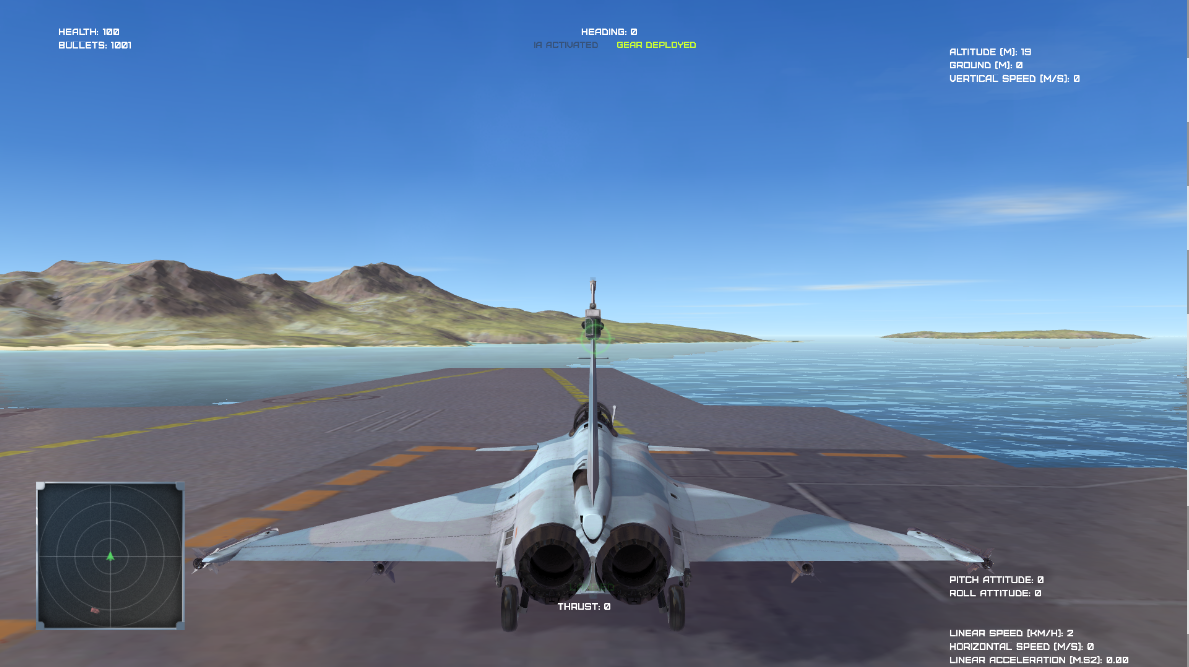}
    \caption{HUD feature provides useful indicators about the flight (altitude, pitch, roll, etc.).}
    \label{fig:ray3}
\end{figure}

\paragraph{2D Overlays Pass}
The HUD and other instructions which describe the various phases of the simulation are shown in 2D, within the screen plane. For instance, HUD feature can be seen in Figure \ref{fig:ray3}.

\paragraph{Post-Processing}
Once the 3D renders and overlays are displayed, a final pass is made which has no other purpose than aesthetics. At this point, it is mainly the fade in/fade out effects that are being applied to provide harmonious transitions between each phase.

\subsubsection{Earth and Space Geometry}

To increase visual realism, the curvature of the horizon and the atmospheric hue change according to the altitude of the aircraft are also simulated. In first person view, in exterior view, or in VR, the immersion is thus reinforced. The essence of the simulation enables us to restrict the altitude that the aircraft can reach thereby avoiding the complexity of a planetary visualization when viewed from orbit. Again, the suggested realism strengthens the acceptance factor of the simulation by the users, without increasing the complexity of the implementation whose primary goal is to be used as a sandbox.

\begin{figure}[ht] 
    \centering
    \includegraphics[scale=0.5]{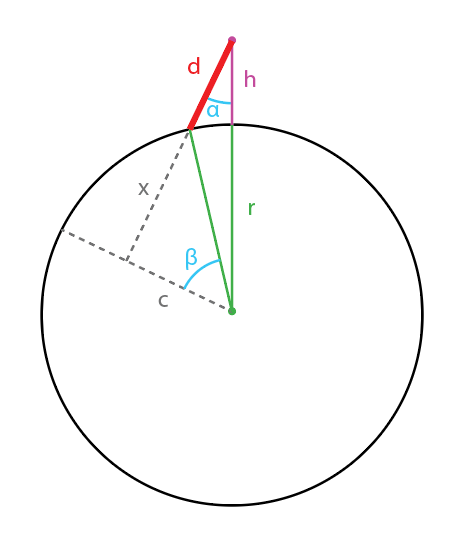}
    \caption{Geometric illustrations of the equations above.}
    \label{fig:geo1}
\end{figure}

\paragraph{Intersection Equation}
We assume that the distance between the camera and the Earth's surface is still very short compared to the planet's radius. We therefore require an equation that gives more accurate results than the standard sphere/ray intersection, by solving a second order equation. You can check the geometric illustrations for this calculations in Figure \ref{fig:geo1}.  

h: camera altitude \\ 
r: Earth's radius \\
$\alpha$: angle between the ray and the surface normal  \\
d: distance on the ray between the camera and the surface\\

Calculation of d: 

$$d=cos(\alpha) \times (r + h) - x$$ 
$$x=r \times sin(\beta)$$  

Calculation of $\beta$:  

$$c=sin(\alpha) \times (r + h) = r \times cos(\beta)$$  

$$cos(\beta) = \dfrac{sin(\alpha) \times (r + h)}{r}$$  

$$\beta=acos \left( \dfrac{sin(\alpha) \times (r + h)}{r}\right)$$  

We then have our distance d along the ray:

$$d=cos(\alpha) \times (r + h) - r \times sin \left( acos \left( \dfrac{sin(\alpha) \times (r + h)}{r}\right) \right)$$

\paragraph{Altitude Accuracy Issue }
The method described in this document suffers from one limitation: at low altitudes, the numerical accuracy is not sufficient. 
If h is about 100 m, the radius of the Earth r is about 6,000,000 meters. The precision of 32-bit floating point numbers doesn't allow such a large range:

Low altitude $h<<r$:  
$$d=\dfrac{h}{cos(\alpha)}$$

\begin{figure}[ht] 
    \centering
    \includegraphics[scale=0.15]{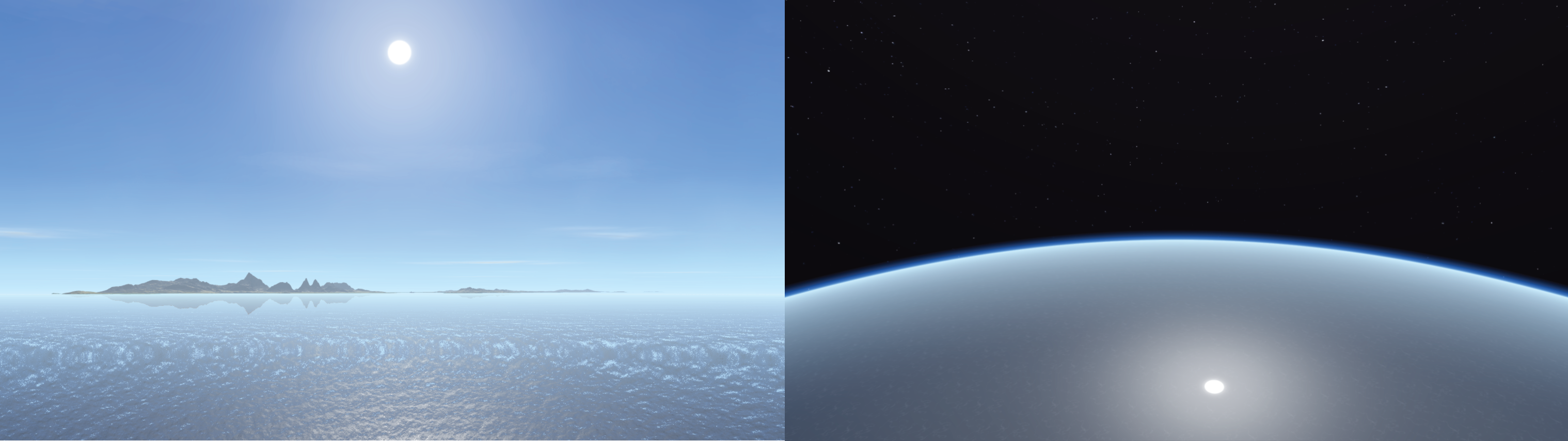}
    \caption{Left image shows the sky view at low altitudes and right image shows the sky view at high altitudes.}
    \label{fig:gokyuzu}
\end{figure}

The answer is to mix the intersection with a plane and the intersection with the sphere. The planar intersection occurs when the magnitude of h is insignificant versus r, whereas the spherical intersection occurs when h grows big enough compared to r. For transitory levels of h relative to r, a fading applies between the two distances.  
The tests showed that the transition is not noticeable to the viewer. Figure \ref{fig:gokyuzu} examplifies a low and a high altitude sky view.

High altitude $h\approx r$:  
$$d=cos(\alpha) \times (r + h) - r \times sin \left( acos \left( \dfrac{sin(\alpha) \times (r + h)}{r}\right) \right)$$  
$$d=\dfrac{h}{cos(\alpha)}$$

\begin{figure}[ht] 
    \centering
    \includegraphics[scale=0.4]{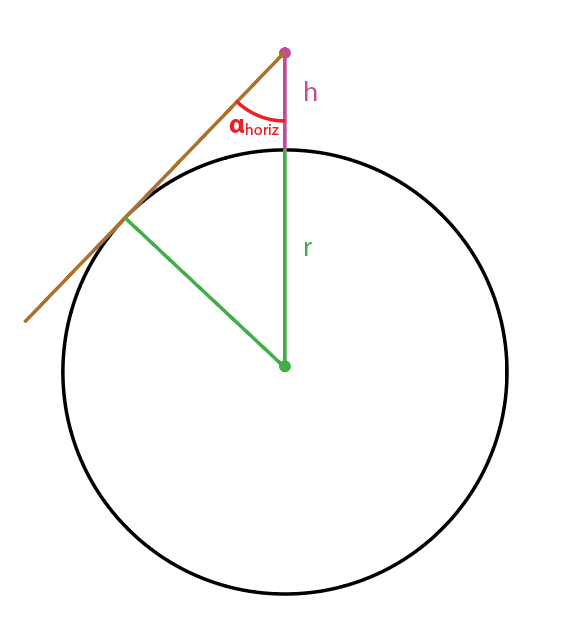}
    \caption{Geometric illustration of the ratio of the camera’s altitude to the Earth’s radius.}
    \label{fig:hesap1}
\end{figure}

\paragraph{Interpolation Between Planar and Spherical Distances}
To compute the interpolation between the planar distance and the spherical distance, we need to figure out the angle between the normal of the surface of the sphere and the horizon line. Geometric illustration of the calculation is shown in Figure \ref{fig:hesap1}.

$$\alpha_{horiz} = \dfrac{\pi}{2} - atan \left( \dfrac {\sqrt{h \times (h + 2 \times r)}}{r} \right)$$  

Then we compute the ratio of the camera's altitude to the Earth's radius:

$$ratio=\dfrac{h}{r}$$
\begin{enumerate}
    \item If ratio < 0.001, then only the planar distance should be used.
    \item If ratio > 0.01, then only the spherical distance is used.
    \item If 0.001< ratio < 0.01, an interpolation applies between the planar distance and the spherical distance, following the angle of the radius.
\end{enumerate}

\paragraph{Generating The Atmosphere}
We define 3 colours:
\begin{enumerate}
    \item Colour of the lower atmosphere (a tone of light blue).
    \item Colour of the upper atmosphere (blue).
    \item Colour of the outer space beyond the atmosphere (black).
\end{enumerate}

Then we will need the angle between the line of the horizon and the edge of the atmosphere $\alpha_{atm}$.  
Two cases are possible:
\begin{enumerate}
    \item The camera is within the atmosphere  
    \item The camera is above the atmosphere  
\end{enumerate}

To determine the colour of the atmosphere, we may choose between a "physical" and an "aesthetic" model.  

The physical model is to estimate the amount of light diffracted and absorbed by the atmosphere. For this purpose, we would need to find out the thickness of the atmosphere crossed by the ray. However, this quickly leads to a complexity that is not necessarily relevant in this case.  

For the Sandbox, the aesthetic approach was preferred, providing greater control over the atmospheric rendering as a function of the aircraft's altitude.

When the camera is inside the atmosphere, we can compute $\alpha_{atm}$ like this:  

$$\alpha_{atm}=\pi - \alpha_{horiz}$$  

However, this poses a special difficulty when the camera is just above the atmosphere:

\begin{figure}[ht] 
    \centering
    \includegraphics[scale=0.6]{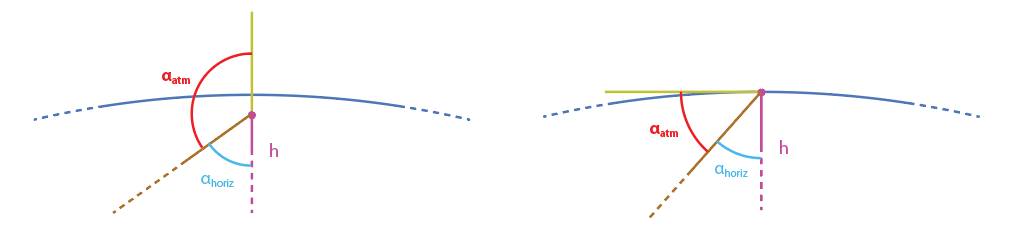}
    \caption{Change in $\alpha_{atm}$ according to h.}
    \label{fig:hesap2}
\end{figure}
 As seen from figure \ref{fig:hesap2} when $h$ goes above the atmosphere,
 $$\alpha_{atm} = \dfrac{\pi}{2} - \alpha_{horiz}$$

To circumvent this sharp edge, we can interpolate the angles along the thickness of the atmosphere.  
To do this, we determine the parametric altitude $F$ of the camera within the atmosphere:
$$F=\dfrac{h}{A_{t}}$$  
\begin{figure}[ht] 
    \centering
    \includegraphics[scale=0.4]{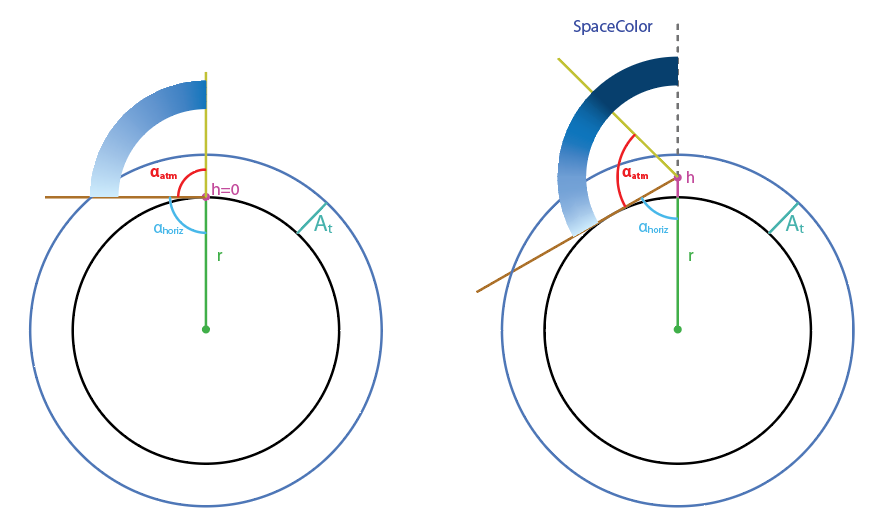}
    \caption{Sky/Space color and viewing angles when the camera is within the atmosphere.}
    \label{fig:ic}
\end{figure}

\paragraph{F <= 1 : Camera Within The Atmosphere}
$$\alpha_{atm} = \pi \times (1-F) + (\dfrac{\pi}{2} \times F) - \alpha_{horiz}$$  \\
An interpolation applies to the colour beyond the atmosphere:  
\begin{enumerate}
    \item F=0 : SpaceColor = colour of the upper atmosphere  
    \item F=1 : SpaceColor = colour of space beyond the atmosphere  
\end{enumerate}

\paragraph{F > 1 : Camera Above The Atmosphere}
In this case, the equation is a variation of the calculation of the angle between the normal of the sphere and the horizon line.

$$\alpha_{atm} = \dfrac{\pi}{2} - atan \left( \dfrac {\sqrt{(h-A_{t}) \times ((h-A_{t}) + 2 \times (r+A_{t}))}}{r+A_{t}} \right) - \alpha_{horiz}$$  

\begin{figure}[ht] 
    \centering
    \includegraphics[scale=0.4]{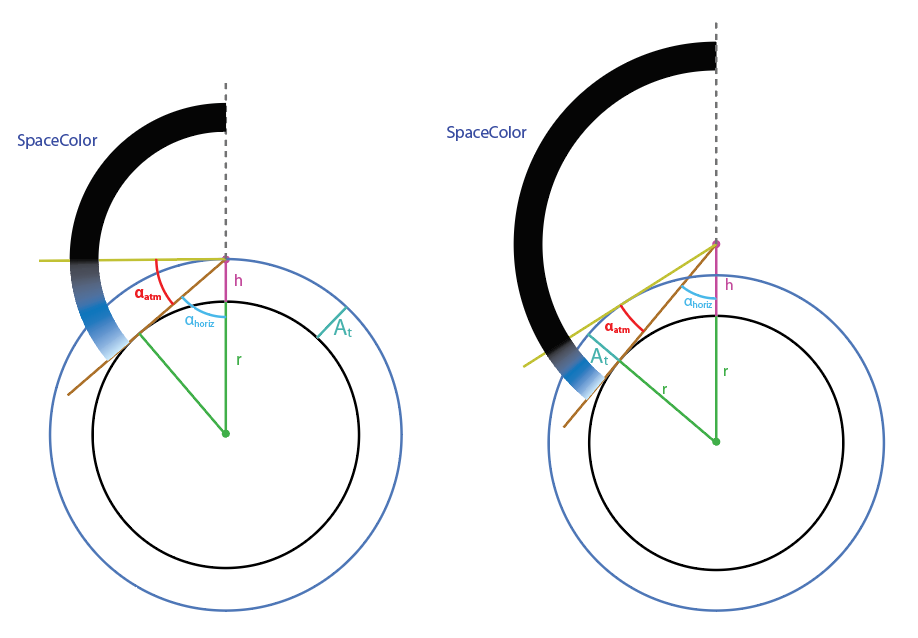}
    \caption{Sky/Space color and viewing angles when the camera is above the atmosphere.}
    \label{fig:dis}
\end{figure}

\subsubsection{Pursuit Camera Model}
The camera is aimed at a position slightly delayed in time from the aircraft. To render the movement more realistically, a subtle Perlin noise is added to the orientation. The intensity of this noise varies according to three parameters: the velocity of the aircraft, the afterburner activation, the acceleration intensity.

\subsection{Renderless Mode}
It is known that 3D rendering with high graphics, puts a lot of stress on the processing units and inhibits the performance of the computer. It also unnecessarily raises the system requirements for training, making it impossible to run on many device. Once the renderless mode is turned on, rendering is deactivated while the algorithm continues learning in the background. Thus, excessive performance constraint is disposed of. Renderless mode also makes running the software on server side possible since the program does not impose any graphical (visual) output.

\section{Experiment}
So far, we have discussed diverse scenarios that can be implemented in Harfang. In this part, for the purpose of validating the usability of our simulation environment, a simple navigation task is experimented. The details of this experiment are discussed below.

\begin{figure}[ht] 
    \centering
    \includegraphics[scale=0.7]{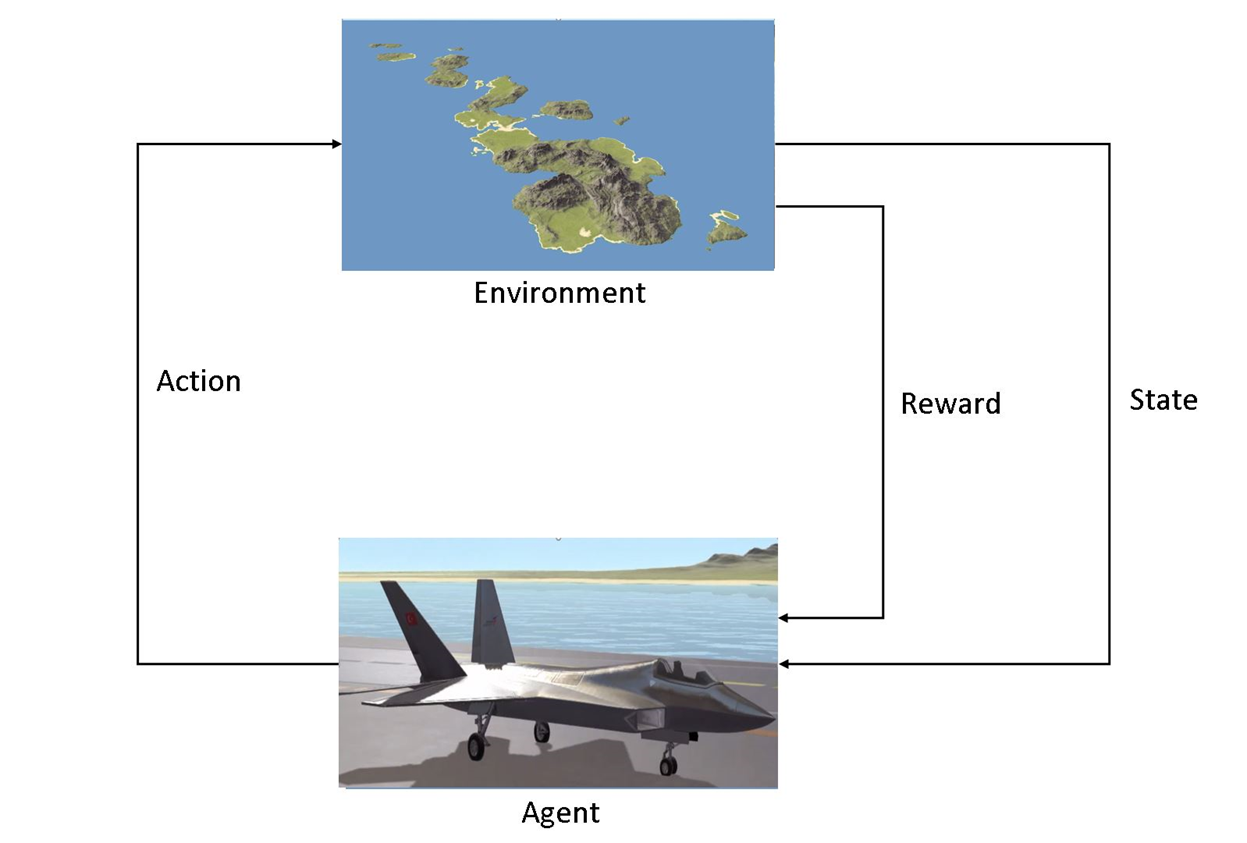}
    \caption{Interaction of a reinforcement learning agent with the environment.}
\end{figure}

\subsection{Scenario}

\begin{figure}[ht] 
    \centering
    \includegraphics[scale=0.7]{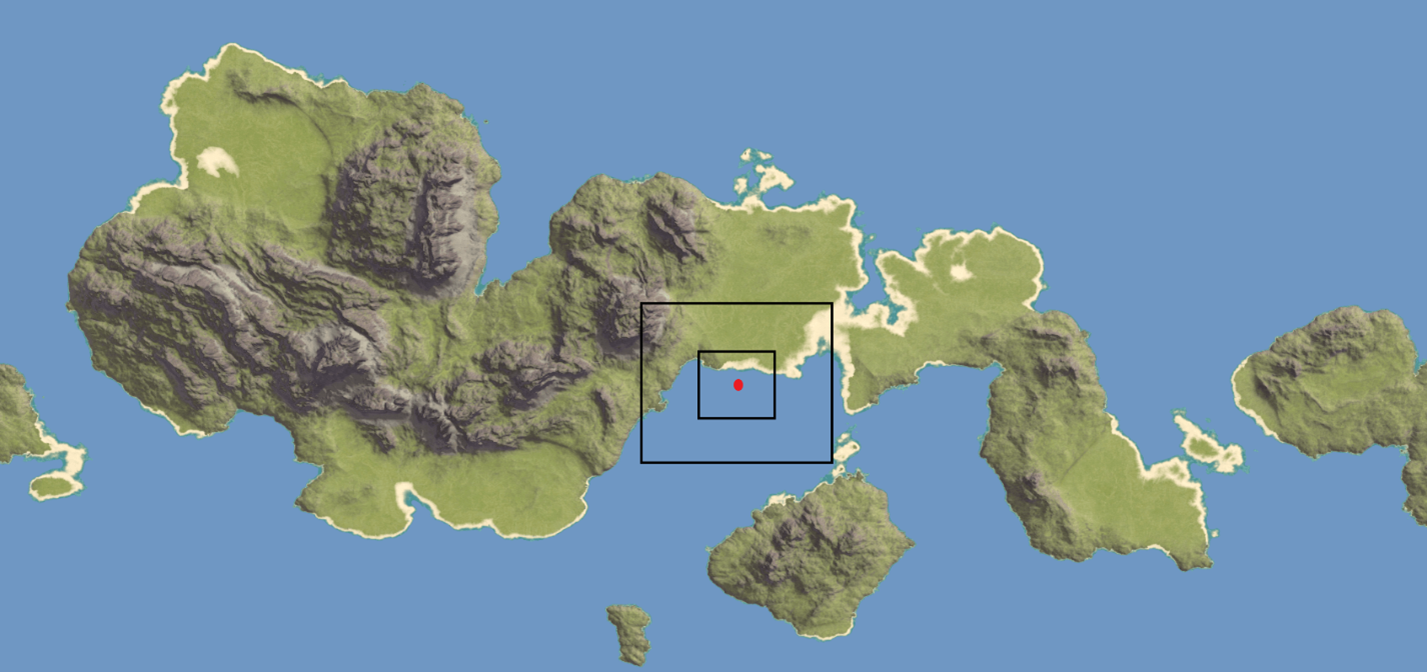}
    \caption{Playground of the agent. The red dot in the center represents the goal and the two black squares are the borders of starting region.}
    \label{fig:playground}
\end{figure}

In this experiment, agents are initialized at a random position and altitude within the region lies between the two black square shown in Figure ~\ref{fig:playground}. The task is to reach the destination point which is marked as red regardless of the starting position.

\subsection{Method}

TD3 + HER are used in this project, TD3 \cite{otuziki} was published in 2018. The main difference of this method when compared to other methods is that even if it uses actor-critic as a base, we cannot deny that classical methods like DDPG \cite{otuzuc} have some problem. Those problems are mainly an overestimation, it basically can be explained as giving actions to agents in places it did not present, so this is the main reason why TD3 + HER come up with better results when compared to DDPG and classical actor-critic methods. We tested DDPG at the beginning, but we did not able to perform even basic actions in DDPG, that is why we have moved to TD3 instead.

\subsection{Action Space}

For this problem, we adopted a low-level control approach. This means that the agent directly manipulates the control surfaces of the plane. Therefore, there were 3 actions outputted by the algorithm: rudder, elevator, aileron. The reason why we choose a low-level control strategy is because we intended to see what the agent is capable of learning without limiting its abilities.

\subsection{State Space}

The agent had access to 13 state signals in total: (x, y, z) coordinate differences between the goal and the agent, $(\phi, \theta, \psi)$ Euler angles, horizontal and vertical velocity, heading angle, pitch attitude, and acceleration in (x, y, z) directions. Although there were other signals present in the simulation environment, we decided that these signals were adequate in order for agent to accomplish the task.

\subsection{Reward}

We tried to keep the reward as simple as possible without hindering the agents ability to learn how to reach the goal. Hence, the reward structured according to the distance of the plane from the goal position as depicted in the formula below. This reward formula aims to promote the agent going towards the goal as it receives higher reward with the decreasing distance because the reward function outputs a negative value. In addition to that, the agent receives +100 reward if it reaches the goal, and -100 if it crashes or cannot reach the goal within the duration of an episode.

\begin{equation}
R = -10^{-5}\sqrt{(UAV_x-Goal_x)^2 + (UAV_y-Gaol_y)^2 + (UAV_z-Goal_z)^2}
\label{eq_ASME}
\end{equation}

\begin{table}[ht]
    \centering
    \caption{Hyperparameter setting used in the experiment.}
    \begin{tabular}{ll}
        \toprule
        Hyper-Parameter                          & Values      \\ \cmidrule{1-2}
        
        Critic Learning Rate             & $10^{-3}$      \\
        Actor Learning Rate     & $10^{-3}$      \\
        Optimizer     & Adam    \\
        Target Update Rate ($\tau$)     & $5 . 10^{-3}$      \\
        Batch Size     & 100    \\
        Iterations per time step     & 1     \\
        Discount Factor     & 0.99      \\
        Normalized Observations     & True     \\
        Gradient Clipping     & False      \\
        Actor First Hidden Layer Neuron number     & 400      \\
        Actor Second Hidden Layer Neuron number     & 300      \\
        Critic First Hidden Layer Neuron number     & 400      \\
        Critic Second Hidden Layer Neuron number     & 300      \\ \bottomrule

    \end{tabular}
\end{table}

\section{Result}
After a training of about 1400 episodes, our agent was capable of navigating to the desired destination. It can be seen from Figure ~\ref{fig:result} that the learning slowed down at about 800 episodes and came to a convergence at 1000 episodes. Even though there were some visible oscillations in the reward before 1000 episodes, at the end, our agent managed to exploit the correct behaviours and successfully learned to perform this basic navigation task. This results proves that Harfang3D Dog-Fight Sandbox is an eligible platform for reinforcement learning research offering a broad array of scenarios, high accessibility, easy implementation, and extremely flexible training opportunities with renderless mode and distributional structure.

\begin{figure}[ht] 
    \centering
    \includegraphics[scale=0.17]{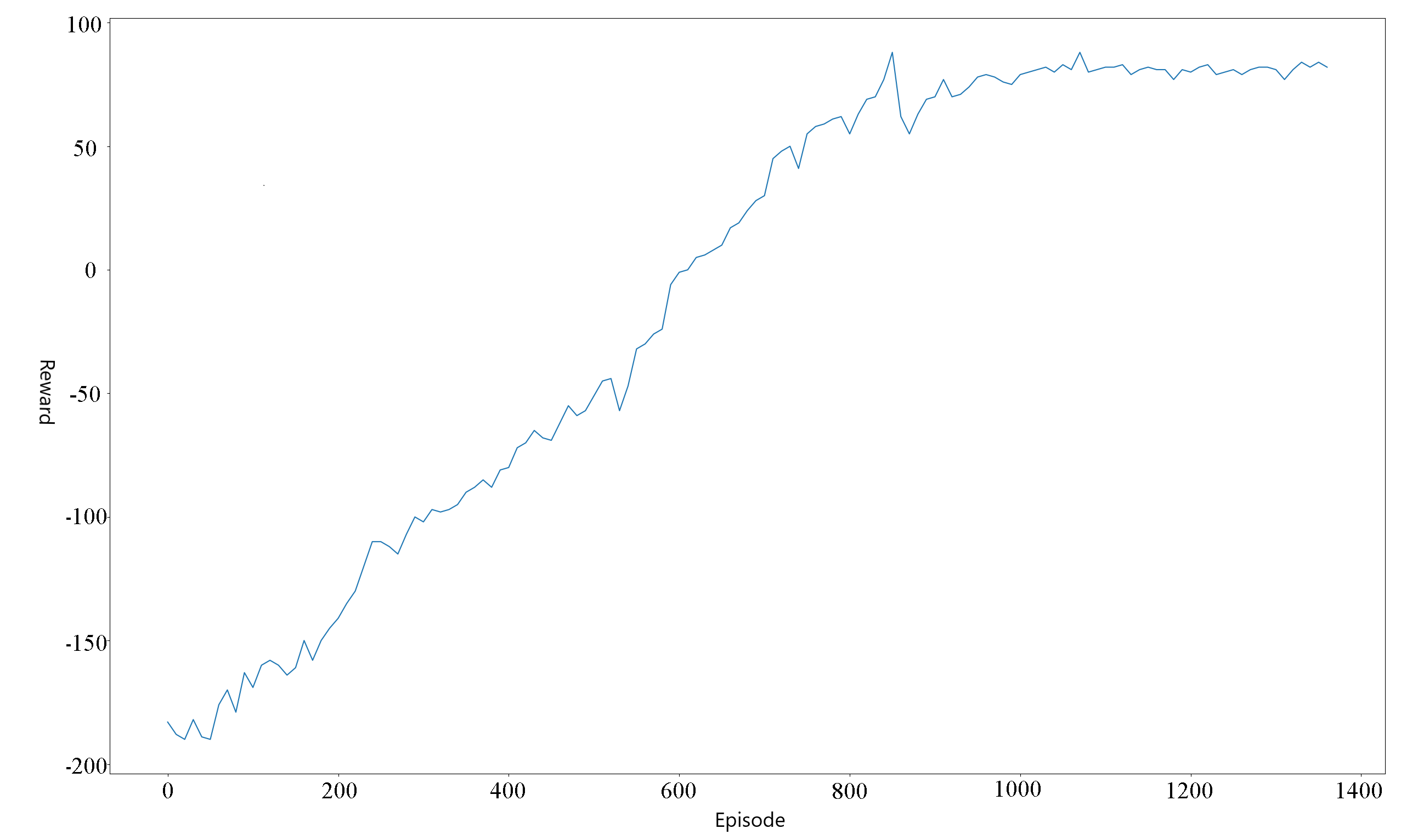}
    \caption{Episodic reward plot of the training phase.}
\end{figure}
\section*{Acknowledgements}
We would like to thank Thomas Simonnet, François Gutherz, Philippe Herber, Serge Bidault and the other members of the HARFANG team for the technical support.


\bibliographystyle{ieeetr}






\end{document}